\pdfoutput=1

\documentclass[11pt]{article}

\usepackage[preprint]{acl}

\usepackage{times}
\usepackage{latexsym}
\usepackage[T1]{fontenc}
\usepackage[utf8]{inputenc}
\usepackage{microtype}
\usepackage{inconsolata}
\usepackage{graphicx}

\usepackage{float} 
\usepackage{graphicx} 
\usepackage{subcaption} 
\usepackage{multirow} 
\usepackage{diagbox} 
\usepackage{color,soul} 
\usepackage{linguex} 
\usepackage{amsmath}
\usepackage{bm} 
\usepackage{arydshln} 
\usepackage{comment} 
\usepackage{xfrac} 
\usepackage{xurl} 

\usepackage{tikz}
\usepackage{scalerel}
\usepackage{pict2e}
\usepackage{tkz-euclide}
\usetikzlibrary{calc}
\usetikzlibrary{patterns,arrows.meta}
\usetikzlibrary{shadows}
\usetikzlibrary{external}
\usepackage{adjustbox}

\definecolor{primaryLight}{HTML}{C0392B}
\definecolor{new_green}{rgb}{0.35, 0.69, 0.19}
\definecolor{new_red}{rgb}{0.85, 0, 0.}
\definecolor{new_blue}{rgb}{0., 0.5, 1.}
\definecolor{new_yellow}{rgb}{0.98, 0.85, 0.37}
\definecolor{new_gray}{rgb}{0.86, 0.86, 0.86}
\definecolor{new_beige_light}{rgb}{0.98, 0.92, 0.84}
\definecolor{new_beige_dark}{rgb}{0.8, 0.58, 0.46}

\title{Multi-Lingual Implicit Discourse Relation Recognition \\ with Multi-Label Hierarchical Learning}

\author{Nelson Filipe Costa \and Leila Kosseim \\
        Computational Linguistics at Concordia (CLaC) Laboratory \\
        Department of Computer Science and Software Engineering \\
        Concordia University, Montr\'eal, Qu\'ebec, Canada \\
        \texttt{nelsonfilipe.costa@mail.concordia.ca} \\
        \texttt{leila.kosseim@concordia.ca}}

\begin{document}

\maketitle


\begin{abstract}

This paper introduces the first multi-lingual and multi-label classification model for implicit discourse relation recognition (IDRR). Our model, HArch, is evaluated on the recently released DiscoGeM~2.0 corpus and leverages hierarchical dependencies between discourse senses to predict probability distributions across all three sense levels in the PDTB~3.0 framework. We compare several pre-trained encoder backbones and find that RoBERTa-HArch achieves the best performance in English, while XLM-RoBERTa-HArch performs best in the multi-lingual setting. In addition, we compare our fine-tuned models against GPT-4o and Llama-4-Maverick using few-shot prompting across all language configurations. Our results show that our fine-tuned models consistently outperform these LLMs, highlighting the advantages of task-specific fine-tuning over prompting in IDRR. Finally, we report SOTA results on the DiscoGeM~1.0 corpus, further validating the effectiveness of our hierarchical approach.

\end{abstract}


\section{Introduction}

Discourse analysis explores how textual segments are connected through relations of coherence. One of the most widely adopted frameworks for computational discourse analysis is the Penn Discourse Treebank (PDTB)~\citep{miltsakaki2004penn,prasad2008penn}. In the PDTB framework, a discourse relation is established between two textual segments (referred to as arguments) either explicitly, through the presence of a discourse connective (e.g., \textit{but} and \textit{because}), or implicitly when no connective is present. The sense of the relation is then classified using a defined list of discourse senses, which are organized hierarchically across three levels~\citep{webber2019penn}. The task of identifying the sense of an implicit relation is known as implicit discourse relation recognition (IDRR).

Research on IDRR has mostly been conducted on the English annotated PDTB corpora~\citep{corpus-pdtb-2,corpus-pdtb-3}. However, the subjective nature of discourse interpretation poses challenges for single-label annotation schemes, where each instance is assigned only one sense label ~\citep{stede2008disambiguation,scholman2017examples, hoek-etal-2021-less}. The ambiguity in the annotation of implicit relations has been further highlighted by the challenges in mapping them across discourse frameworks~\citep{demberg2019compatible,costa2023mapping}. As a response, recent works have advocated for multi-label annotation, recognizing that multiple discourse senses can often co-occur within a single relation~\citep{yung-etal-2019-crowdsourcing,pyatkin-etal-2020-qadiscourse,scholman-etal-2022-discogem,scholman-etal-2022-design,pyatkin2023design,yung-etal-2024-discogem,yung-demberg-2025-crowdsourcing}. This shift in perspective has led to the development of the multi-label annotated DiscoGeM~1.0 corpus~\citep{scholman-etal-2022-discogem} for implicit discourse relations.

This perspective shift toward multi-label annotation has stirred an initial wave of research in multi-label IDRR~\citep{costa2024exploring, long2024multi, costa2024multi}. However, as with the PDTB corpora, the DiscoGeM~1.0 corpus is limited to the English language and most work in the field of discourse continues to focus mostly on the English language. While single-label discourse annotated corpora following the PDTB framework exist in other languages, such as the German Potsdam Commentary Corpus~2.2~\citep{bourgonje-stede-2020-potsdam} and the Czech Prague Discourse Treebank~3.0~\citep{synkova-etal-2024-announcing}, multi-lingual corpora remain scarce. For instance, the TED-MDB corpus~\citep{zeyrek-etal-2024-multiple} provides parallel annotations of single-label discourse relations across seven languages. However, its size remains relatively small for the training of IDRR models - it contains $1,774$ implicit relations. To address this gap and enable cross-linguistic research in IDRR, the recently released DiscoGeM~2.0 corpus~\citep{yung-etal-2024-discogem} introduces parallel multi-label annotations of implicit discourse relations following the PDTB~3.0 sense hierarchy in four languages: English, German, French and Czech.

In this paper, we leverage the DiscoGeM~2.0 corpus to present the first multi-lingual and multi-label classification model for IDRR. Our main contributions are as follows:

\begin{itemize}
    \item We propose a hierarchical multi-task architecture, HArch, that leverages dependencies across all three sense levels in the PDTB~3.0 framework, improving on prior multi-label approaches~\citep{costa2024multi}.

    \item We conduct the first evaluation of multi-label IDRR on the DiscoGeM~2.0 corpus in the English, German, French and Czech languages individually and in a multi-lingual setting.

    \item We compare different pre-trained encoder backbones and find that RoBERTa-HArch and XLM-RoBERTa-HArch achieve the best results in the English-only and multi-lingual settings, respectively.

    \item We evaluate our fine-tuned models against LLMs, including GPT-4o and Llama-4-Maverick, using few-shot prompting and show that our models consistently outperform them in multi-label IDRR.

    \item We benchmark our best-performing model on the English-only DiscoGeM~1.0 corpus and achieve new SOTA results.
\end{itemize}


\section{Previous Work}

The advent of transformer-based models~\citep{vaswani2017attention} has led to substantial progress in natural language understanding. Nevertheless, IDRR remains one of the most challenging tasks in computational discourse analysis. Traditionally formulated as a single-label classification problem, most approaches have addressed IDRR in English either by fine-tuning~\citep{long-webber-2022-facilitating,liu-strube-2023-annotation} or prompt-tuning~\citep{chan-etal-2023-DiscoPrompt,zhao-etal-2023-infusing,zeng-etal-2024-global,long2024leveraging} pre-trained language models (PLMs). Another line of research has also explored directly prompting large language models (LLMs) through prompt-engineering to solve single-label IDRR~\citep{chan-etal-2024-exploring,yung-etal-2024-prompting}. However, results from both work indicate that zero-shot and few-shot prompting of LLMs continues to significantly underperform when compared to the results obtained through fine-tuning and prompt-tuning PLMs. The same outcome has been observed in other IDRR work focusing on different languages~\citep{saeed-etal-2025-implicit,ruby-etal-2025-multimodal}.

The recent move toward multi-label annotation in discourse data has led to initial efforts in multi-label IDRR. \citet{long2024multi} used the 4.9\% of implicit discourse relations in the PDTB~3.0 corpus that are annotated with two senses to build a model capable of predicting up to two senses per instance. However, given that most of PDTB~3.0 annotations remain single-label, their model predominantly produces single-label predictions. In contrast, \citet{yung-etal-2022-label} and \citet{costa2024exploring} used the multi-label DiscoGeM~1.0 corpus but convert its annotations into a single-label format during training. Finally, \citet{costa2024multi} present the first results in multi-label IDRR by training a model on DiscoGeM~1.0 that jointly predicts sense distributions across all three levels in the PDTB~3.0 framework. Rather than adapting multi-label data to a single-label setting or modifying a single-label corpus for multi-label use, they preserve the inherent multi-label structure of DiscoGeM~1.0 throughout training. However, like most previous work in IDRR, their model is limited to the English language.

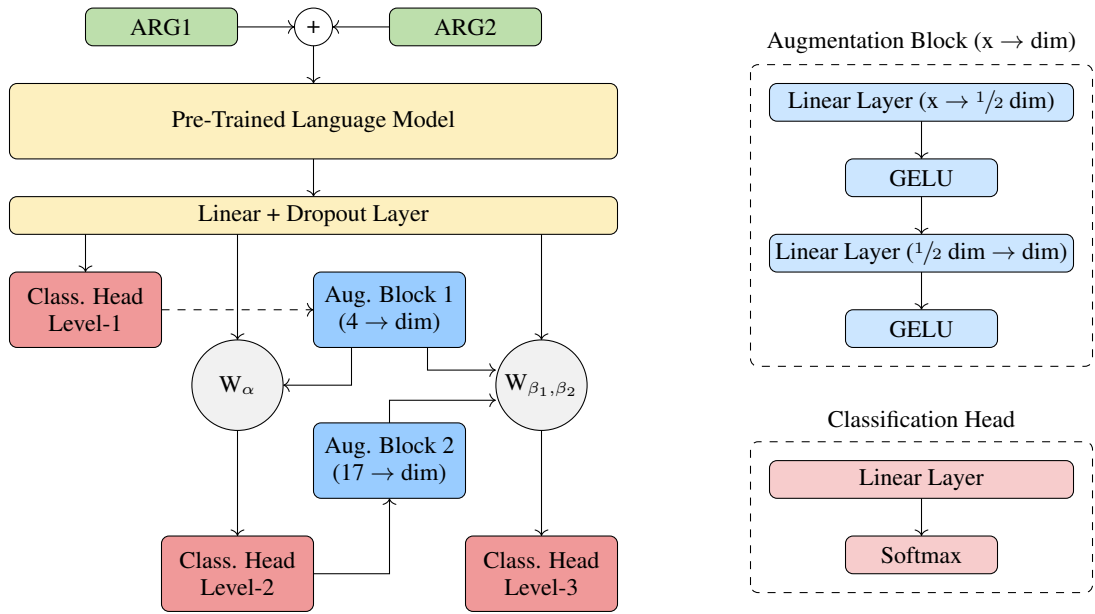
\begin{figure*}
    \centering
    \begin{tikzpicture}[every node/.style={font=\small}]

        \draw[fill=new_green!40, rounded corners=0.1cm] (1,7.5) rectangle (3,8) node[pos=0.5, anchor=center, align=center] {ARG1};

        \draw[fill=new_green!40, rounded corners=0.1cm] (5,7.5) rectangle (7,8) node[pos=0.5, anchor=center, align=center] {ARG2};

        \draw[->] (3,7.75) -- (3.75,7.75);

        \draw[->] (5,7.75) -- (4.25,7.75);

        \draw (4,7.75) circle (0.25) node {+};

        \draw [->] (4,7.5) -- (4,7);

        \draw[fill=new_yellow!40, rounded corners=0.1cm] (0,6) rectangle (8,7) node[pos=0.5, anchor=center, align=center] {Pre-Trained Language Model};

        \draw [->] (4,6) -- (4,5.5);

        \draw[fill=new_yellow!40, rounded corners=0.1cm] (0,5) rectangle (8,5.5) node[pos=0.5, anchor=center, align=center] {Linear + Dropout Layer};

        \draw [->] (1,5) -- (1,4.5);

        \draw [->] (3,5) -- (3,3.6);

        \draw [->] (7,5) -- (7,3.6);

        \draw[fill=new_red!40, rounded corners=0.1cm] (0,3.5) rectangle (2,4.5) node[pos=0.5, anchor=center, align=center] {Class. Head \\ Level-1};

        \draw [->, dashed] (2,4) -- (4,4);

        \draw[fill=new_blue!40, rounded corners=0.1cm] (4,3.5) rectangle (6,4.5) node[pos=0.5, anchor=center, align=center] {Aug. Block 1 \\ (4 $\rightarrow$ dim)};

        \draw [->] (4.5,3.5) -- (4.5,3) -- (3.6,3);

        \draw [->] (5.5,3.5) -- (5.5,3.2) -- (6.4,3.2);

        \draw[fill=new_gray!40] (3,3) circle (0.6) node {W$_\alpha$};

        \draw [->] (3,2.4) -- (3,1);

        \draw[fill=new_gray!40] (7,3) circle (0.6) node {W$_{\beta_1,\beta_2}$};

        \draw [->] (7,2.4) -- (7,1);

        \draw[fill=new_blue!40, rounded corners=0.1cm] (4,1.5) rectangle (6,2.5) node[pos=0.5, anchor=center, align=center] {Aug. Block 2 \\ (17 $\rightarrow$ dim)};

        \draw [->] (5,2.5) -- (5,2.8) -- (6.4,2.8);

        \draw[fill=new_red!40, rounded corners=0.1cm] (2,0) rectangle (4,1) node[pos=0.5, anchor=center, align=center] {Class. Head \\ Level-2};

        \draw [->] (4,0.5) -- (5,0.5) -- (5,1.5);

        \draw[fill=new_red!40, rounded corners=0.1cm] (6,0) rectangle (8,1) node[pos=0.5, anchor=center, align=center] {Class. Head \\ Level-3};

        \node at (12, 7.55) {Augmentation Block (x $\rightarrow$ dim)};

        \draw[rounded corners=0.1cm, dashed] (9.75,7.25) rectangle (14.25,3.25);

        \draw[fill=new_blue!20, rounded corners=0.1cm] (10,6.5) rectangle (14,7) node[pos=0.5, anchor=center, align=center] {Linear Layer (x $\rightarrow$ $\sfrac{1}{2}$ dim)};

        \draw [->] (12,6.5) -- (12,6);

        \draw[fill=new_blue!20, rounded corners=0.1cm] (11,5.5) rectangle (13,6) node[pos=0.5, anchor=center, align=center] {GELU};

        \draw [->] (12,5.5) -- (12,5);

        \draw[fill=new_blue!20, rounded corners=0.1cm] (10,4.5) rectangle (14,5) node[pos=0.5, anchor=center, align=center] {Linear Layer ($\sfrac{1}{2}$ dim $\rightarrow$ dim)};

        \draw [->] (12,4.5) -- (12,4);

        \draw[fill=new_blue!20, rounded corners=0.1cm] (11,3.5) rectangle (13,4) node[pos=0.5, anchor=center, align=center] {GELU};

        \node at (12, 2.55) {Classification Head};

        \draw[rounded corners=0.1cm, dashed] (9.75,0.25) rectangle (14.25,2.25);

        \draw[fill=new_red!20, rounded corners=0.1cm] (10,1.5) rectangle (14,2) node[pos=0.5, anchor=center, align=center] {Linear Layer};

        \draw [->] (12,1.5) -- (12,1);

        \draw[fill=new_red!20, rounded corners=0.1cm] (11,0.5) rectangle (13,1) node[pos=0.5, anchor=center, align=center] {Softmax};
    \end{tikzpicture}
    \caption{Architecture of our hierarchical multi-task model, HArch, for multi-label IDRR. The model takes as input a concatenated pair of discourse arguments and generates probability distributions across all three sense levels in the PDTB~3.0. The output of the lower-level classification heads is projected onto the embedding space of the encoder via augmentation blocks and then combined with the shared representation using learnable weighted sums. This architecture explicitly models hierarchical dependencies and enables joint learning across all sense levels.}
    \label{fig:model}
\end{figure*}


\section{Our Approach}
\label{sec:approach}

We propose a hierarchical multi-task classification model, HArch, to predict probability distributions across all three sense levels in the PDTB~3.0 framework. We also propose a prompting framework to compare our HArch model against SOTA LLMs through few-shot learning.


\subsection{HArch Model}
\label{sec:harch}

Our HArch model improves on the model of \citet{costa2024multi} by explicitly modeling hierarchical dependencies across sense levels. Figure~\ref{fig:model} shows the architecture of our model. The input of the model is a concatenated pair of discourse arguments (ARG1 + ARG2), which is encoded using a pre-trained language model. The resulting contextual representation is fed to a shared linear transformation followed by a dropout layer. This shared representation is then passed to three separate classification heads - one for each sense level. The level-1 classification head predicts a probability distribution over the 4 possible level-1 senses using a linear layer followed by a softmax activation - as shown in red on the bottom-right side of Figure~\ref{fig:model}. This 4-dimensional output vector is then projected onto the same embedding space of the pre-trained language model through an augmentation block. As shown in blue on the upper-right side of Figure~\ref{fig:model}, this augmentation is done in two steps: the output of the classification head is first mapped to a hidden dimension of half the size of the embedding space and then, in a second step, further projected to match the full embedding size. A GELU~\citep{hendrycks2016gaussian} activation function after each incremental step introduces non-linearity, while helping with sparsity and preventing exploding activations in the dimensionality projection. Based on early experiments, this incremental dimensional projection ensured a smoother transition in feature space. The augmented level-1 output is then combined with the shared representation through a weighted sum defined by Equation~\ref{eq:alpha}, where $\alpha$ is a learnable parameter of the model, to yield the input of the level-2 classification head.

\begin{equation}
    \mathbf{W}_\alpha = \alpha \cdot \mathbf{h}_{\text{Aug}_1} + (1 - \alpha) \cdot \mathbf{h}_{\text{Dropout}}
    \label{eq:alpha}
\end{equation}

The level-2 classification head then predicts a probability distribution over the 17 possible level-2 senses. Similar to level-1, the level-2 output is passed through a dimensionality augmentation block to project it into the same embedding space of the pre-trained language model. The augmented level-1 and level-2 outputs are then added to the shared representation through a second weighted sum defined by Equation~\ref{eq:betas}, where $\beta_1$ and $\beta_2$ are learnable parameters. The output of this weighted sum is then fed into the level-3 classification head, which predicts a probability distribution over the level-3 senses.

\begin{equation}
    \begin{split}
        \mathbf{W}_{\beta_1,\beta_2} = & \beta_1 \cdot \mathbf{h}_{\text{Aug}_1} + \beta_2 \cdot \mathbf{h}_{\text{Aug}_2} + \\
        & (1 - \beta_1 - \beta_2) \cdot \mathbf{h}_{\text{Dropout}}
    \end{split}
    \label{eq:betas}
\end{equation}

This cascading of information ensures that the predictions of lower-level classification heads are used to help inform the prediction of higher-level classification heads that require more fine-grained information. Finally, we use the Adam optimization algorithm~\cite{diederik2015adam} to minimize the loss of the model, which we calculate as the sum of the losses of each classification head. We use the Mean Absolute Error function to calculate the loss of each classification head as it was shown to lead to better performance in in multi-label classification~\citep{costa2024multi}. To evaluate our model, we use Jensen-Shannon (JS) distance to measure the similarity between the predicted and the reference probability distributions, similarly to other multi-label classification works in NLP~\citep{pyatkin2023design,yung-etal-2024-discogem,van2024annotator,costa2024multi}.


\subsection{Few-Shot Prompting}
\label{sec:prompt}

Despite prior work showing poor results in IDRR through prompt engineering~\citep{chan-etal-2024-exploring,yung-etal-2024-prompting}, we compared our HArch model to LLMs in multi-label IDRR via direct prompting with few-shot learning. We tested GPT-4o~\citep{hurst2024gpt} and Llama-4-Maverick with the prompt structure described in Appendix~\ref{apx:prompt}. Depending on the language evaluated, the prompt was translated and examples were drawn from the corresponding language subset of DiscoGeM~2.0. When considering all languages simultaneously in a multi-lingual setting, we kept the structure of the prompt in English and provided examples in the different languages. All of the examples were taken from the training split of the corpus. To ensure consistency, we set the temperature to $0$ and allowed up to $5$ retries per instance in cases where prompt outputs did not conform to the expected format. Prompting costs were approximately $5.00\$$ for GPT-4o and $0.19-0.49\$$ for Llama-4-Maverick per million input tokens.

The prompt was designed to closely replicate the annotation methodology of DiscoGeM~2.0. When crowdsourcing the annotation of the corpus, the authors provided the annotators a list of non-ambiguous connectives~\citep{yung-etal-2024-discogem} that represented each of the possible senses in the PDTB~3.0 framework. This list of connectives was adapted for each language. The annotators were then asked to choose the connective that best expresses the semantic relation between the arguments of the relation being annotated, irrespective of whether syntax would need to be adjusted. By mirroring this annotation process in our prompting template, we were able to directly evaluate the performance of both LLMs on multi-label IDRR using the test split of DiscoGeM~2.0.


\section{Data Preparation}
\label{sec:data}

To train our model on multi-lingual IDRR, we use the recently released DiscoGeM~2.0 corpus~\citep{yung-etal-2024-discogem}. The corpus comprises a total of 13,063\footnote{In their paper they report a total of 12,834 implicit discoutse relations, out of which 5,618 are in English. However, we counted 5,847 discourse relations in English in the corpus.} parallel annotated implicit discourse relations across four languages: 5,847 in English, 2,588 in German, 2,628 in French, and 2,000 in Czech. All annotations follow the PDTB~3.0 discourse framework~\citep{webber2019penn}. While its predecessor, DiscoGeM~1.0~\cite{scholman-etal-2022-discogem}, provided the first corpus of multi-label implicit discourse relations, DiscoGeM~2.0 is the first corpus to provide multi-lingual parallel annotation of multi-label implicit discourse relations.

Each implicit discourse relation in the corpus was annotated by at least 10 crowdworkers. Each annotator selected the sense they considered most appropriate for each relation through a proxy task and then all of the annotated senses of each relation were averaged to produce a multi-label sense distribution over 28 different senses in the PDTB~3.0 framework - the \textsc{Belief} and the \textsc{Speech-Act} senses were not annotated. For each language, the authors of the corpus prepared a list of discourse connectives (one representing each of the available senses) and asked the annotators to select for each relation the discourse connective that best expresses the semantic relation between its two arguments. From the chosen connectives, the authors were able to infer the possible multiple senses of each implicit discourse relation.

\begin{table*}[h]
    \centering
    \renewcommand{\arraystretch}{1.2}
    \scalebox{1}{
        \begin{tabular}{|c|r|r|r|r|r|}
            \hline
            \hline
            Level-1 & \multicolumn{1}{c|}{English} & \multicolumn{1}{c|}{German} & \multicolumn{1}{c|}{French} & \multicolumn{1}{c|}{Czech} & \multicolumn{1}{c|}{All} \\
            \hline
            \hline
            Temporal & $556.8$ & $438.4$ & $400.7$ & $458.8$ & $1,854.6$ \\
            Contingency & $1,695.4$ & $772.3$ & $811.0$ & $681.1$ & $3,959.8$ \\
            Comparison & $796.3$ & $346.5$ & $525.0$ & $227.1$ & $1,894.9$ \\
            Expansion & $2,798.6$ & $1,030.8$ & $891.3$ & $632.9$ & $5,353.7$ \\
            \hline
            \hline
            Total & $5,847.1$ & $2,588.0$ & $2,628.0$ & $1,999.9$ & $13,063$ \\
            \hline
            \hline
        \end{tabular}
    }
    \caption{Distribution of level-1 senses per language in the DiscoGeM~2.0 corpus. Each value represents the sum of the corresponding sense in all the multi-label distributions of implicit discourse relations in the specific language.}
    \label{tab:statistics-l1}
\end{table*}

\begin{table*}[h]
    \centering
    \renewcommand{\arraystretch}{1.2}
    \resizebox{\textwidth}{!}{
        \begin{tabular}{|cc|c|ccc|c|}
            \hline
            \hline
            \multicolumn{2}{|c|}{Language} & \multirow{2}{*}{Model} & \multirow{2}{*}{Level-1} & \multirow{2}{*}{Level-2} & \multirow{2}{*}{Level-3} & \multirow{2}{*}{Params} \\
            \cline{1-2}
            Test & Train &  &  &  &  &  \\
            \hline
            \hline
            \multirow{6}{*}{Eng} & \multirow{6}{*}{Eng} & \citet{costa2024multi}
                                             & $0.329 \pm 0.005$ & $0.498 \pm 0.004$ & $0.569 \pm 0.005$ & 125M \\
             &  & RoBERTa-HArch                 & $\bm{0.327 \pm 0.004}$ & $\bm{0.478 \pm 0.005}$ & $\bm{0.541 \pm 0.005}$ & 125M \\
             &  & ModernBERT-HArch              & $0.364 \pm 0.004$ & $0.507 \pm 0.005$ & $0.592 \pm 0.006$ & 149M \\
             &  & XLM-RoBERTa-HArch             & $0.343 \pm 0.004$ & $0.487 \pm 0.006$ & $0.564 \pm 0.005$ & 125M \\
             &  & \cellcolor{new_gray!40} GPT-4o & \cellcolor{new_gray!40} $0.373 \pm 0.002$ & \cellcolor{new_gray!40} $0.559 \pm 0.003$ & \cellcolor{new_gray!40} $0.657 \pm 0.003$ & \cellcolor{new_gray!40} 200B \\
             &  & \cellcolor{new_gray!40} Llama-4-Maverick & \cellcolor{new_gray!40} $0.423 \pm 0.004$ & \cellcolor{new_gray!40} $0.641 \pm 0.006$ & \cellcolor{new_gray!40} $0.711 \pm 0.007$ & \cellcolor{new_gray!40} 400B \\
            \hdashline
            \multirow{5}{*}{All} & \multirow{5}{*}{All} & XLM-RoBERTa-HArch & $\bm{0.356 \pm 0.005}$ & $\bm{0.531 \pm 0.006}$ & $\bm{0.605 \pm 0.006}$ & 125M \\
             &  & EuroBERT-HArch                        & $0.414 \pm 0.006$ & $0.569 \pm 0.007$ & $0.641 \pm 0.006$ & 210M \\
             &  & Flan-T5-HArch                         & $0.364 \pm 0.003$ & $0.550 \pm 0.003$ & $0.623 \pm 0.004$ & 220M \\
             &  & \cellcolor{new_gray!40} GPT-4o & \cellcolor{new_gray!40} $0.438 \pm 0.002$ & \cellcolor{new_gray!40} $0.671 \pm 0.002$ & \cellcolor{new_gray!40} $0.709 \pm 0.003$ & \cellcolor{new_gray!40} 200B \\
             &  & \cellcolor{new_gray!40} Llama-4-Maverick & \cellcolor{new_gray!40} $0.534 \pm 0.004$ & \cellcolor{new_gray!40} $0.774 \pm 0.004$ & \cellcolor{new_gray!40} $0.839 \pm 0.005$ & \cellcolor{new_gray!40} 400B \\
            \hline
            \hline
        \end{tabular}
    }
    \caption{Experimental results using HArch with different PLM encoders. The results are the average JS distance of three different runs on the test split of DiscoGeM 2.0. Results are reported using the English language and all languages simultaneously. Lower scores indicate better performance. Values in bold show the best score at each level in each language setting. Rows shaded in gray present the results of LLMs with few-shot prompting. The final column reports the approximate number of parameters of each model.}
    \label{tab:model-selection}
\end{table*}

To ensure reproducibility of our results, we split the DiscoGeM~2.0 corpus following the proposed training, validation and test splits~\citep{yung-etal-2024-discogem}. Contrary to the common approach in traditional single-label IDRR, we follow the methodology of \citet{costa2024multi} and consider all three sense levels in the PDTB~3.0 framework when possible. Table~\ref{tab:statistics-l1} shows the distribution of level-1 senses and Table~\ref{tab:statistics-l2-l3} in Appendix~\ref{apx:statistics} shows the distribution of level-2 and level-3 senses per language in the DiscoGeM~2.0 corpus. We replaced the non-existent PDTB~3.0 level-3 senses with their corresponding level-2 sense.


\section{Encoder Selection}
\label{sec:experiments}

In this section, we compare the performance of different PLMs as encoders in our HArch model (see Figure~\ref{fig:model}) for multi-label IDRR in both multi-lingual and English-only settings, using the DiscoGeM~2.0 corpus. Table~\ref{tab:model-selection} summarizes the performance of each configuration on the validation split of the corpus. Each model was trained for 10 epochs with a batch size of 16 and the results are reported as the average Jensen-Shannon (JS) distance across three independent runs. In addition to experimenting with different PLM encoders, we also report results obtained using LLMs via few-shot prompting. All of the code and prompt templates are available on GitHub\footnote{\url{https://github.com/nelsonfilipecosta/Multi-Lingual-Implicit-Discourse-Relation-Recognition}}.

To evaluate the impact of incorporating hierarchical learning in multi-label IDRR, we first compare our proposed HArch model against the non-hierarchical model proposed by~\citet{costa2024multi}. For a fair comparison, both models use the same pre-trained RoBERTa$_{\text{base}}$ encoder~\citep{liu2019roberta} and are trained under identical conditions with a learning rate of $1e^{-5}$. To our knowledge, \citet{costa2024multi} is the only existing approach modeling multi-label IDRR with probability distributions over the full set of PDTB~3.0 sense labels. Since their model was only trained in the English language, we compared the performance of our models using only the English subset of DiscoGeM~2.0. As shown in Table~\ref{tab:model-selection}, our model RoBERTa-HArch, which incorporates hierarchical dependencies across sense levels, achieves lower JS distances at level-2 and level-3 compared to the non-hierarchical model of \citet{costa2024multi}. At level-1, the performance of both models is comparable, which is to be expected as level-1 predictions do not benefit from lower-level inputs within the hierarchical structure.

\begin{table*}[h]
    \centering
    \renewcommand{\arraystretch}{1.2}
    \scalebox{1}{
        \begin{tabular}{|c|ccc|}
            \hline
            \hline
            Model & Level-1 & Level-2 & Level-3 \\
            \hline
            \hline
            \citet{costa2024multi} & $\bm{0.299 \pm 0.002}$ & $0.446 \pm 0.003$ & $0.523 \pm 0.002$ \\
            \hdashline
            RoBERTa-HArch          & $0.302 \pm 0.003$ & $\bm{0.435 \pm 0.003}$ & $\bm{0.506 \pm 0.004}$ \\
            \hline
            \hline
        \end{tabular}
    }
    \caption{Final results showing the average JS distance of three different runs for each model on the test split of DiscoGeM~1.0. Lower scores indicate better performance. Values in bold show the best score at each level.}
    \label{tab:discogem-1}
\end{table*}

Next, we explored the impact of using different PLMs as encoder backbones for our HArch hierarchical model. In particular, we experimented with ModernBERT$_{\text{base}}$~\citep{warner2024smarter}, a recently introduced encoder claimed to outperform RoBERTa, and the multi-lingual XLM-RoBERTa$_{\text{base}}$~\citep{conneau-etal-2020-unsupervised}. Contrary to expectations, ModernBERT-HArch consistently underperformed across all three sense levels, as shown in Table~\ref{tab:model-selection}. Even XLM-RoBERTa-HArch, despite its multi-lingual design, achieved lower JS distances than ModernBERT-HArch - though, it still fell short of the performance achieved by RoBERTa-HArch. In addition to fine-tuned models, we also evaluated LLMs via few-shot prompting. As shown in Table~\ref{tab:model-selection}, GPT-4o substantially outperforms Llama-4-Maverick in terms of JS distance across all levels, replicating a pattern already observed between GPT-3.5 and LLaMa~\citep{touvron2023llama} in a similar setting~\citep{yung-etal-2024-prompting}. However, despite its stronger performance in a few-shot setting, GPT-4o still does not surpass our smaller fine-tuned RoBERTa-HArch model, which remains the best-performing system overall in this multi-label IDRR task.

In the multi-lingual setting, we evaluated PLMs pre-trained on multiple languages as encoder backbones for HArch. Specifically, we experimented with XLM-RoBERTa$_{\text{base}}$, FLAN-T5$_{\text{base}}$~\citep{chung2024scaling} and the recently released EuroBERT$_{\text{210m}}$~\citep{boizard2025eurobert}. We also adapted the prompt template described in Appendix~\ref{apx:prompt} to include examples in all four languages covered by DiscoGeM~2.0. Given the relatively limited number of annotated instances in German, French, and Czech within the DiscoGeM~2.0 corpus (see Table~\ref{tab:statistics-l1}), we did not experiment with language specific PLMs for these languages. Instead, we infer model performance in these languages from the results in the multi-lingual setting. As shown in Table~\ref{tab:model-selection}, using XLM-RoBERTa as the encoder backbone of HArch achieves the best overall performance in the multi-lingual configuration, even outperforming the more recent EuroBERT encoder model. Finally, we observe a surprising drop in performance in the LLM models when prompted in the multi-lingual setting compared to when prompted in the English language alone.


\section{Results}
\label{sec:results}

In this section, we report the performance of our HArch models on the test splits of both DiscoGeM~1.0 (see Table~\ref{tab:discogem-1}) and DiscoGeM~2.0 (see Table~\ref{tab:discogem-2}) in a multi-lingual setting, where all languages are used simultaneously, and separately for each language. These evaluations allow us to assess the generalizability of our models across the different languages of the corpus.

We report results on DiscoGeM~1.0 to benchmark our approach with the only other existing model for multi-label IDRR that predicts probability distributions over the full set of PDTB~3.0 sense labels~\citep{costa2024multi}. Since DiscoGeM~1.0 contains only English data, we report the performance of our best-performing English model in Table~\ref{tab:model-selection} - RoBERTa-HArch. As shown in Table~\ref{tab:discogem-1}, RoBERTa-HArch achieves lower JS distances at level-2 and level-3 compared to the non-hierarchical model of \citet{costa2024multi}, while maintaining comparable performance at level-1. These results provide further empirical support for our hypothesis that incorporating hierarchical dependencies improves the ability of the model to generate more accurate probability distributions at finer-grained sense levels.

\begin{table*}[h]
    \centering
    \renewcommand{\arraystretch}{1.2}
    \scalebox{1}{
        \begin{tabular}{|cc|c|ccc|}
            \hline
            \hline
            \multicolumn{2}{|c|}{Language} & \multirow{2}{*}{Model} & \multirow{2}{*}{Level-1} & \multirow{2}{*}{Level-2} & \multirow{2}{*}{Level-3} \\
            \cline{1-2}
            Test & Train &  &  &  &  \\
            \hline
            \hline
            \multirow{3}{*}{All} & All & XLM-RoBERTa-HArch & $\bm{0.353 \pm 0.005}$ & $\bm{0.529 \pm 0.005}$ & $\bm{0.606 \pm 0.04}$ \\
             & $-$ & \cellcolor{new_gray!40} GPT-4o & \cellcolor{new_gray!40} $0.434 \pm 0.002$ & \cellcolor{new_gray!40} $0.678 \pm 0.003$ & \cellcolor{new_gray!40} $0.705 \pm 0.003$ \\
             & $-$ & \cellcolor{new_gray!40} Llama-4-Maverick & \cellcolor{new_gray!40} $0.529 \pm 0.004$ & \cellcolor{new_gray!40} $0.769 \pm 0.005$ & \cellcolor{new_gray!40} $0.830 \pm 0.005$ \\
            \hdashline
            \multirow{5}{*}{Eng} & Eng & RoBERTa-HArch & $\bm{0.331 \pm 0.004}$ & $\bm{0.477 \pm 0.004}$ & $\bm{0.548 \pm 0.005}$ \\
             & All & \multirow{2}{*}{XLM-RoBERTa-HArch} & $0.349 \pm 0.004$ & $0.493 \pm 0.006$ & $0.568 \pm0.005 $ \\
             & Eng &  & $0.347 \pm 0.005$ & $0.486 \pm 0.006$ & $0.561 \pm 0.006$ \\
             & $-$ & \cellcolor{new_gray!40} GPT-4o & \cellcolor{new_gray!40} $0.368 \pm 0.001$ & \cellcolor{new_gray!40} $0.551 \pm 0.001$ & \cellcolor{new_gray!40} $0.647 \pm 0.002$ \\
             & $-$ & \cellcolor{new_gray!40} Llama-4-Maverick & \cellcolor{new_gray!40} $0.420 \pm 0.001$ & \cellcolor{new_gray!40} $0.634 \pm 0.002$ & \cellcolor{new_gray!40} $0.702 \pm 0.002$ \\
            \hdashline
            \multirow{4}{*}{Ger} & All & \multirow{2}{*}{XLM-RoBERTa-HArch} & $\bm{0.360 \pm 0.007}$ & $\bm{0.566 \pm 0.006}$ & $\bm{0.630 \pm 0.006}$ \\
             & Ger &  & $0.393 \pm 0.006$ & $0.586 \pm 0.006$ & $0.656 \pm 0.007$ \\
             & $-$ & \cellcolor{new_gray!40} GPT-4o & \cellcolor{new_gray!40} $0.411 \pm 0.002$ & \cellcolor{new_gray!40} $0.644 \pm 0.002$ & \cellcolor{new_gray!40} $0.725 \pm 0.002$ \\
             & $-$ & \cellcolor{new_gray!40} Llama-4-Maverick & \cellcolor{new_gray!40} $0.509 \pm 0.003$ & \cellcolor{new_gray!40} $0.732 \pm 0.006$ & \cellcolor{new_gray!40} $0.809 \pm 0.007$ \\
            \hdashline
            \multirow{4}{*}{Fre} & All & \multirow{2}{*}{XLM-RoBERTa-HArch} & $\bm{0.367 \pm 0.006}$ & $\bm{0.553 \pm 0.005}$ & $\bm{0.628 \pm 0.006}$ \\
             & Fre &  & $0.382 \pm 0.004$ & $0.576 \pm 0.006$ & $0.655 \pm 0.007$ \\
             & $-$ & \cellcolor{new_gray!40} GPT-4o & \cellcolor{new_gray!40} $0.406 \pm 0.001$ & \cellcolor{new_gray!40} $0.611 \pm 0.001$ & \cellcolor{new_gray!40} $0.689 \pm 0.002$ \\
             & $-$ & \cellcolor{new_gray!40} Llama-4-Maverick & \cellcolor{new_gray!40} $0.473 \pm 0.003$ & \cellcolor{new_gray!40} $0.667 \pm 0.003$ & \cellcolor{new_gray!40} $0.748 \pm 0.004$ \\
            \hdashline
            \multirow{4}{*}{Cze} & All & \multirow{2}{*}{XLM-RoBERTa-HArch} & $\bm{0.371 \pm 0.010}$ & $\bm{0.555 \pm 0.009}$ & $\bm{0.647 \pm 0.009}$ \\
             & Cze &  & $0.404 \pm 0.009$ & $0.584 \pm 0.011$ & $0.689 \pm 0.010$ \\
             & $-$ & \cellcolor{new_gray!40} GPT-4o & \cellcolor{new_gray!40} $0.549 \pm 0.011$ & \cellcolor{new_gray!40} $0.820 \pm 0.010$ & \cellcolor{new_gray!40} $0.854 \pm 0.012$ \\
             & $-$ & \cellcolor{new_gray!40} Llama-4-Maverick & \cellcolor{new_gray!40} $0.613 \pm 0.005$ & \cellcolor{new_gray!40} $0.864 \pm 0.004$ & \cellcolor{new_gray!40} $0.890 \pm 0.005$ \\
            \hline
            \hline
        \end{tabular}
    }
    \caption{Final results showing the average JS distance of three different runs for each model on the test split of DiscoGeM~2.0. Results are reported using all languages simultaneously and each language individually. Lower scores indicate better performance. Values in bold show the best score at each level in each tested language setting. Rows shaded in gray present the results of LLMs with few-shot prompting.}
    \label{tab:discogem-2}
\end{table*}

Table~\ref{tab:discogem-2} presents the results of our models across different language configurations. Since XLM-RoBERTa-HArch achieved the best overall performance in the multi-lingual setting with the validation split of DiscoGeM~2.0 (see Table~\ref{tab:model-selection}), we report its results on the test split of the corpus when trained both on all languages and individually on each language. For English, we also include results from RoBERTa-HArch, which was fine-tuned exclusively on English data alone. In addition, we evaluate GPT-4o and Llama-4-Maverick using language-specific versions of the prompt detailed in Appendix~\ref{apx:prompt}. As shown in Table~\ref{tab:discogem-2}, our HArch models consistently outperform both LLMs across all languages and all sense levels. The performance gap is particularly pronounced in all settings with the exception of the English language. Notably, XLM-RoBERTa-HArch achieves substantially lower JS distances than both LLMs at level-2 and level-3 in the multi-lingual and Czech language. Consistent with earlier results in Table~\ref{tab:model-selection}, GPT-4o outperforms Llama-4-Maverick across all configurations, but still lags behind our smaller fine-tuned HArch models. These findings reinforce the idea that fine-tuning smaller encoder-based models can still outperform prompting large language models in specific tasks~\citep{bucher2024fine,chan-etal-2024-exploring,yung-etal-2024-prompting}.

Focusing on our fine-tuned models, Table~\ref{tab:discogem-2} shows that for German, French and Czech, XLM-RoBERTa-HArch achieves lower JS distances when trained on all languages compared to when trained on the test language alone. This suggests that multi-lingual training helps compensate for the limited annotated data available in these languages. In contrast, for English, XLM-RoBERTa-HArch performs slightly better when fine-tuned exclusively on the English subset of DiscoGeM~2.0. However, in both cases, it fails to outperform RoBERTa-HArch - which remains the strongest model for English overall. When comparing the performance of RoBERTa-HArch on DiscoGeM~1.0 in Table~\ref{tab:discogem-1} and DiscoGeM~2.0 in Table~\ref{tab:discogem-2}, we observe a drop across all sense levels on the latter. This discrepancy can be attributed to the reduced label set used for evaluation on DiscoGeM~1.0. To allow a fair comparison with~\citet{costa2024multi}, we adopted the same adapted set of 14 level-2 sense labels proposed by~\citet{kim-etal-2020-implicit}. This reduced set of level-2, and consequently level-3, sense labels lowers the complexity of the multi-label IDRR task on DiscoGeM~1.0 when compared to DiscoGeM~2.0, where we considered all sense labels on the corpus.


\section{Multi-Task Contribution}
\label{sec:ablation}

\begin{table*}[h]
    \centering
    \renewcommand{\arraystretch}{1.2}
    \resizebox{\textwidth}{!}{
        \begin{tabular}{|cc|c|ccc|c|}
            \hline
            \hline
            \multicolumn{2}{|c|}{Language} & \multirow{2}{*}{Model} & \multirow{2}{*}{Level-1} & \multirow{2}{*}{Level-2} & \multirow{2}{*}{Level-3} & \multirow{2}{*}{Time} \\
            \cline{1-2}
            Test & Train &  &  &  &  &  \\
            \hline
            \hline
            \multirow{2}{*}{Eng} & \multirow{2}{*}{Eng} & RoBERTa-HArch & $\bm{0.331 \pm 0.004}$ & $\bm{0.477 \pm 0.004}$ & $\bm{0.548 \pm 0.005}$ & 9h 09m \\
             &  & RoBERTa-Individual & $0.342 \pm 0.004$ & $0.513 \pm 0.004$ & $0.589 \pm 0.005$ & 21h 56m \\
            \hdashline
            \multirow{2}{*}{All} & \multirow{2}{*}{All} & XLM-RoBERTa-HArch & $\bm{0.353 \pm 0.005}$ & $\bm{0.529 \pm 0.005}$ & $\bm{0.606 \pm 0.04}$ & 20h 31m \\
             &  & XLM-RoBERTa-Individual & $0.365 \pm 0.004$ & $0.552 \pm 0.005$ & $0.641 \pm 0.004$ & 59h 14m \\
            \hline
            \hline
        \end{tabular}
    }
    \caption{Results showing the contribution of multi-task learning. The results are the average JS distance of three different runs on the test split of DiscoGeM~2.0. Results are reported using the English language and all languages simultaneously. Lower scores indicate better performance. Values in bold show the best score at each level in each language setting. The final column reports the total time required to compute all of the averaged results in each row.}
    \label{tab:ablation}
\end{table*}

To evaluate the benefit of jointly modeling all three sense levels in a multi-task setting, we conducted an ablation study by training three individual models - each dedicated to predicting probability distributions for a single sense level. This study was conducted in the English and multi-lingual setting. To ensure a fair comparison, each individual model tries to replicate the architecture of the multi-task model (see Figure~\ref{fig:model}), with the exception that it includes only one classification head and no augmentation blocks. Specifically, each individual model encodes a concatenated pair of discourse arguments using the corresponding PLM, which it then passes to a linear transformation and dropout layer, before feeding the representation to a classification head — similar to the level-1 prediction path in the multi-task model. Table~\ref{tab:ablation} compares the results of the three individual models for each sense level against RoBERTa-HArch in the English setting and XLM-RoBERTa-HArch in the multi-lingual setting, using the test split of DiscoGeM~2.0.

As shown in Table~\ref{tab:ablation}, the HArch model trained in a multi-task setting consistently outperforms the single-level models across all three sense levels. The performance gains are especially pronounced at level-2 and level-3, suggesting that these finer-grained senses benefit significantly from the shared representations and hierarchical cues provided by joint training. While level-1 predictions also improve slightly, the main advantage of multi-task learning emerges from its ability to propagate information downward through the sense hierarchy. This ablation study isolates the impact of our architecture and confirms that jointly modeling the hierarchy enhances multi-label IDRR performance. In addition to yielding better performance, HArch is also more efficient - as shown by the total time it took to calculate the averaged results in each row of Table~\ref{tab:ablation}  with a 32-core compute node with 512GB of RAM. This highlights both the effectiveness and efficiency of our proposed architecture.


\section{Conclusion}

In this work, we presented the first multi-lingual and multi-label classification model for IDRR. We modeled hierarchical dependencies between discourse senses to predict probability distributions across all three sense levels in the PDTB~3.0 framework. We introduced the first evaluation on the DiscoGeM~2.0 corpus across all languages and compared different pre-trained language models as encoder backbones within our proposed HArch model architecture. Results show that RoBERTa-HArch achieves the best performance in English, while XLM-RoBERTa-HArch performs best in the multi-lingual setting. We further evaluated our models against GPT-4o and Llama-4-Maverick using few-shot prompting across all language configurations and found that the HArch models consistently obtained lower JS distances. These findings demonstrate that fine-tuning smaller encoder-based models can outperform prompting large language models in the context of multi-label IDRR. Lastly, we benchmarked RoBERTa-HArch on the English-only DiscoGeM~1.0 corpus and achieved SOTA results compared to existing work.

These findings not only advance the field of IDRR, but also highlight the advantages of fine-tuning smaller encoder-based models over prompting large language models — particularly, given the significantly higher monetary and environmental costs associated with the latter. Looking ahead, we plan to investigate how discourse sense predictions vary across languages in DiscoGeM~2.0 for parallel discourse relations. In particular, we are interested in exploring how translation influences discourse reasoning and label distribution across these four languages. Additionally, we aim to compare the performance of using XLM-RoBERTa as the encoder in our HArch model in the German, French and Czech languages against encoders pre-trained specifically in these languages. This line of research may shed light on cross-linguistic challenges in discourse analysis and lead to more robust multi-lingual IDRR models.


\section{Limitations}

The PDTB~3.0 framework defines a total of 22 level-2 senses and 36 level-3 senses (after projection from level-2 senses). However, the \textsc{Belief} and \textsc{SPEECH-ACT} senses were not annotated in the DiscoGeM~2.0 corpus. As a result, our HArch model could only be trained on 17 level-2 and 28 level-3 senses (after projection from level-2 senses). Should these additional senses be annotated in future versions of the dataset, our model could be readily extended to incorporate them.

Another limitation of this work concerns the choice of PLMs used to generate embeddings for the pair of discourse arguments. Due to computational and time constraints, we did not experiment with larger models such as LLaMA~3 \citep{grattafiori2024llama}, which might have led to improved performance. Although our experiments were conducted using a high-performance computing infrastructure, we relied on relatively smaller models, such as RoBERTa, to reduce resource demands. Nevertheless, the environmental impact of such experiments remains non-negligible. Future work should take into account the energy consumption associated with fine-tuning PLMs and prompting LLMs and consider it as an additional metric when evaluating model performance.


\section*{Acknowledgements}

The authors would like to thank the anonymous reviewers for their comments. This work was financially supported by the Natural Sciences and Engineering Research Council of Canada (NSERC).


\bibliography{nelson}


\appendix


\section{Data Statistics}
\label{apx:statistics}

Table~\ref{tab:statistics-l2-l3} shows the distribution of level-2 and level-3 senses per language in the DiscoGeM~2.0 corpus. We replaced the non-existent PDTB~3.0 level-3 senses with their corresponding level-2 sense and marked them with * in the \textit{Level-3} column. Each value represents the sum of the corresponding sense in all the multi-label distributions of implicit discourse relations in the specific language subset.

\begin{table*}
    \centering
    \renewcommand{\arraystretch}{1.4}
    \resizebox{\textwidth}{!}{
        \begin{tabular}{|c|r|r|r|r|r|c|r|r|r|r|r|}
            \hline
            \hline
            Level-2 & \multicolumn{1}{c|}{English} & \multicolumn{1}{c|}{German} & \multicolumn{1}{c|}{French} & \multicolumn{1}{c|}{Czech} & \multicolumn{1}{c|}{All} & Level-3 & \multicolumn{1}{c|}{English} & \multicolumn{1}{c|}{German} & \multicolumn{1}{c|}{French} & \multicolumn{1}{c|}{Czech} & \multicolumn{1}{c|}{All} \\
            \hline
            \hline
            \textsc{Synchronous} & $99.4$ & $191.2$ & $139.8$ & $254.6$ & $685.0$ & \textsc{Synchronous*} & $99.4$ & $191.2$ & $139.8$ & $254.6$ & $685.0$ \\
            \hline
            \multirow{2}{*}{\textsc{Asynchronous}} & \multirow{2}{*}{$457.4$} & \multirow{2}{*}{$247.1$} & \multirow{2}{*}{$260.9$} & \multirow{2}{*}{$204.2$} & \multirow{2}{*}{$1169.6$} & \textsc{Precedence} & $420.7$ & $222.7$ & $240.0$ & $193.7$ & $1,077.1$ \\
            \cdashline{7-12}
             &  &  &  &  &  & \textsc{Succession} & $36.7$ & $24.4$ & $20.9$ & $10.5$ & $92.5$ \\
            \hline
            \multirow{2}{*}{\textsc{Cause}} & \multirow{2}{*}{$1,690.7$} & \multirow{2}{*}{$457.5$} & \multirow{2}{*}{$608.6$} & \multirow{2}{*}{$576.7$} & \multirow{2}{*}{$3,333.5$} & \textsc{Reason} & $386.9$ & $243.9$ & $237.6$ & $281.3$ & $1,149.8$ \\
            \cdashline{7-12}
             &  &  &  &  &  & \textsc{Result} & $1,303.8$ & $213.6$ & $371.0$ & $295.4$ & $2,183.7$ \\
            \cdashline{7-12}
            \hline
            \multirow{2}{*}{\textsc{Condition}} & \multirow{2}{*}{$1.2$} & \multirow{2}{*}{$126.1$} & \multirow{2}{*}{$93.2$} & \multirow{2}{*}{$42.0$} & \multirow{2}{*}{$262.5$} & \textsc{ARG1-as-Cond} & $0.1$ & $109.4$ & $87.9$ & $25.1$ & $222.4$ \\
            \cdashline{7-12}
              &  &  &  &  &  & \textsc{ARG2-as-Cond} & $1.1$ & $16.7$ & $5.4$ & $16.9$ & $40.1$ \\
            \hline
            \multirow{2}{*}{\textsc{Neg-Condition}} & \multirow{2}{*}{$1.1$} & \multirow{2}{*}{$14.0$} & \multirow{2}{*}{$14.0$} & \multirow{2}{*}{$12.9$} & \multirow{2}{*}{$42.0$} & \textsc{ARG1-as-NegCond} & $1.0$ & $7.8$ & $10.9$ & $10.9$ & $30.6$ \\
            \cdashline{7-12}
             &  &  &  &  &  & \textsc{ARG2-as-NegCond} & $0.1$ & $6.2$ & $3.1$ & $2.0$ & $11.5$ \\
            \hline
            \multirow{2}{*}{\textsc{Purpose}} & \multirow{2}{*}{$2.3$} & \multirow{2}{*}{$174.8$} & \multirow{2}{*}{$95.2$} & \multirow{2}{*}{$49.5$} & \multirow{2}{*}{$321.8$} & \textsc{ARG1-as-Goal} & $1.4$ & $114.8$ & $91.6$ & $46.5$ & $254.2$\\
            \cdashline{7-12}
             &  &  &  &  &  & \textsc{ARG2-as-Goal} & $0.9$ & $60.0$ & $3.6$ & $3.1$ & $67.6$\\
            \hline
            \multirow{2}{*}{\textsc{Concession}} & \multirow{2}{*}{$505.1$} & \multirow{2}{*}{$161.3$} & \multirow{2}{*}{$261.3$} & \multirow{2}{*}{$160.9$} & \multirow{2}{*}{$1088.6$} & \textsc{ARG1-as-Denier} & $167.5$ & $61.9$ & $91.3$ & $39.9$ & $360.5$ \\
            \cdashline{7-12}
             &  &  &  &  &  & \textsc{ARG2-as-Denier} & $337.7$ & $99.4$ & $169.9$ & $121.0$ & $728.1$ \\
            \hline
            \textsc{Contrast} & $187.4$ & $98.7$ & $173.3$ & $40.8$ & $500.2$ & \textsc{Contrast*} & $187.4$ & $98.7$ & $173.3$ & $40.8$ & $500.2$ \\
            \hline
            \textsc{Similarity} & $103.7$ & $86.5$ & $90.4$ & $25.5$ & $306.1$ & \textsc{Similarity*} & $103.7$ & $86.5$ & $90.4$ & $25.5$ & $306.1$ \\
            \hline
            \textsc{Conjunction} & $1,298.8$ & $260.0$ & $121.8$ & $155.9$ & $1,836.5$ & \textsc{Conjunction*} & $1,298.8$ & $260.0$ & $121.8$ & $155.9$ & $1,836.5$ \\
            \hline
            \textsc{Disjunction} & $2.7$ & $6.5$ & $3.0$ & $3.1$ & $15.3$ & \textsc{Disjunction*} & $2.7$ & $6.5$ & $3.0$ & $3.1$ & $15.3$ \\
            \hline
            \textsc{Equivalence} & $19.0$ & $154.7$ & $101.5$ & $57.4$ & $332.6$ & \textsc{Equivalence*} & $19.0$ & $154.7$ & $101.5$ & $57.4$ & $332.6$ \\
            \hline
            \multirow{2}{*}{\textsc{Exception}} & \multirow{2}{*}{$1.8$} & \multirow{2}{*}{$26.8$} & \multirow{2}{*}{$41.0$} & \multirow{2}{*}{$6.8$} & \multirow{2}{*}{$76.4$} & \textsc{ARG1-as-Exception} & $0.3$ & $14.7$ & $30.7$ & $2.8$ & $48.5$ \\
            \cdashline{7-12}
             &  &  &  &  &  & \textsc{ARG2-as-Exception} & $1.4$ & $12.2$ & $10.3$ & $4.0$ & $27.8$ \\
            \hline
            \multirow{2}{*}{\textsc{Instantiation}} & \multirow{2}{*}{$352.4$} & \multirow{2}{*}{$99.3$} & \multirow{2}{*}{$124.2$} & \multirow{2}{*}{$84.4$} & \multirow{2}{*}{$660.3$} & \textsc{ARG1-as-Instance} & $16.5$ & $0.0$ & $35.6$ & $25.9$ & $78.0$ \\
            \cdashline{7-12}
             &  &  &  &  &  & \textsc{ARG2-as-Instance} & $335.9$ & $99.3$ & $88.6$ & $58.5$ & $582.3$ \\
            \hline
            \multirow{2}{*}{\textsc{Level-of-Detail}} & \multirow{2}{*}{$1079.1$} & \multirow{2}{*}{$393.2$} & \multirow{2}{*}{$363.1$} & \multirow{2}{*}{$278.8$} & \multirow{2}{*}{$2114.1$} & \textsc{ARG1-as-Detail} & $154.9$ & $88.7$ & $51.8$ & $110.1$ & $405.6$ \\
            \cdashline{7-12}
             &  &  &  &  &  & \textsc{ARG2-as-Detail} & $924.1$ & $304.5$ & $311.3$ & $168.6$ & $1708.5$ \\
            \hline
            \multirow{2}{*}{\textsc{Manner}} & \multirow{2}{*}{$4.5$} & \multirow{2}{*}{$33.7$} & \multirow{2}{*}{$124.0$} & \multirow{2}{*}{$37.2$} & \multirow{2}{*}{$199.4$} & \textsc{ARG1-as-Manner} & $1.3$ & $26.6$ & $112.3$ & $22.7$ & $163.0$ \\
            \cdashline{7-12}
             &  &  &  &  &  & \textsc{ARG2-as-Manner} & $3.1$ & $7.0$ & $11.7$ & $14.6$ & $36.5$ \\
            \hline
            \multirow{2}{*}{\textsc{Substitution}} & \multirow{2}{*}{$40.5$} & \multirow{2}{*}{$56.6$} & \multirow{2}{*}{$12.7$} & \multirow{2}{*}{$9.3$} & \multirow{2}{*}{$119.1$} & \textsc{ARG1-as-Substitution} & $0.0$ & $7.6$ & $6.0$ & $1.3$ & $14.9$ \\
            \cdashline{7-12}
             &  &  &  &  &  & \textsc{ARG2-as-Substitution} & $40.5$ & $49.0$ & $6.7$ & $8.0$ & $104.1$ \\
            \hline
            \hline
        \end{tabular}
    }
    \caption{Distribution of level-2 and level-3 senses per language in the DiscoGeM~2.0 corpus. Each value represents the sum of the corresponding sense in all the multi-label distributions of implicit discourse relations in the specific language subset. Senses at level-3 marked with * represent level-2 senses that were projected to level-3 since they do not exist in the PDTB~3.0.}
    \label{tab:statistics-l2-l3}
\end{table*}


\section{Prompt Template}
\label{apx:prompt}

Figure~\ref{fig:prompt} illustrates the prompt-response template used to evaluate GPT-4o and LLaMA-4-Maverick on multi-label IDRR using the English subset of DiscoGeM~2.0. For evaluations in other languages, we translated the prompt template and selected examples from the corresponding language subset of the corpus. In the multi-lingual setting, where all four languages are used simultaneously, we kept the prompt structure in English and provided examples in all four languages. Due to space constraints, Figure~\ref{fig:prompt} only shows two examples of discourse relations in the initial \textit{User Message} prompt. However, we included five examples when prompting the models. All of the examples were taken from the training split of DiscoGeM~2.0.

We designed the prompt to closely replicate the annotation methodology of DiscoGeM~2.0. When crowdsourcing the annotation of the corpus, the authors provided the annotators a list of non-ambiguous connectives~\citep{yung-etal-2024-discogem} that represented each of the possible senses in the PDTB~3.0 framework. This list of connectives was adapted for each language. The annotators were then asked to choose the connective that best expresses the semantic relation between the arguments of the relation being annotated, irrespective of whether syntax would need to be adjusted. By mirroring this annotation process in our prompting template, we were able to directly evaluate the performance of both LLMs on multi-label IDRR using the test split of DiscoGeM~2.0.

\begin{figure*}
    \centering
    \begin{tikzpicture}[every node/.style={font=\small}]

        \draw[rounded corners=0.1cm] (0,0) rectangle (15,23);

        \fill[new_gray!40] (0,22.9) arc[start angle=180, end angle=90, radius=0.1cm] -- (14.9,23) arc[start angle=90, end angle=0, radius=0.1cm] -- (15,22) -- (0,22);

        \node[anchor=north west] at (0.5, 22.75) {\textbf{Prompt template in English}};

        \draw (0,22) -- (15,22);

        \fill[new_gray!0] (0,22) rectangle (15,16.75);

        \node[anchor=north west] at (0.5, 21.75) {\textbf{System Message:}};

        \node[anchor=north west, text width=13.75cm, align=left] at (0.5, 21.25) {You're an expert linguist with the task of identifying the best discourse connectives to fill in the gap between two arguments. Below you have the ordered list of 28 discourse connectives you can use.\\[0.2cm]

        ["at the same time", "then", "after", "because", "as a result", "in that case", "if", "if not", "unless", "for that purpose", "so that", "even though", "nonetheless", "on the other hand", "similarly", "also", "or", "in other words", "other than that", "an exception is that", "this illustrates that", "for example", "in short", "in more detail", "thereby", "as if", "rather than", "instead"]\\[0.2cm]

        For each pair of arguments given to you, you have to choose the connectives from this list that could be used to connect the two arguments. Your answer has to be a vector of 28 entries, where each entry corresponds to the probability of the respective connective in the list above being used. The sum of all probabilities in the vector must be equal to 1.};

        \draw (0,16.75) -- (15,16.75);

        \fill[new_gray!40] (0,16.75) rectangle (15,10);

        \node[anchor=north west] at (0.5, 16.5) {\textbf{User Message:}};

        \node[anchor=north west, text width=13.75cm, align=left] at (0.5, 16) {Here are a few examples.\\[0.2cm]
        
        Argument 1: There, comrades, is the answer to all our problems. It is summed up in a single word - Man.\\
        Argument 2: Man is the only real enemy we have. Remove Man from the scene, and the root cause of hunger and overwork is abolished for ever.\\[0.2cm]

        Answer: [0.0, 0.1, 0.0,	0.3, 0.0, 0.0, 0.0, 0.0, 0.0, 0.0, 0.0, 0.0, 0.0, 0.0, 0.0, 0.0, 0.0, 0.1, 0.0, 0.0, 0.0, 0.0, 0.3, 0.2, 0.0, 0.0, 0.0, 0.0]\\[0.2cm]
        
        ... (x3)\\[0.2cm]

        Argument 1: It had been a good one. There had been a flying motorcycle in it.\\
        Argument 2: He had a funny feeling he'd had the same dream before. His aunt was back outside the door.\\[0.2cm]

        Answer: [0.0, 0.0, 0.0, 0.0, 0.111, 0.0, 0.0, 0.0, 0.0, 0.0, 0.0, 0.0, 0.111, 0.0, 0.0, 0.667, 0.0, 0.0, 0.0, 0.0, 0.0, 0.0, 0.0, 0.111, 0.0, 0.0, 0.0, 0.0]\\
        };

        \draw (0,10) -- (15,10);

        \fill[new_gray!0] (0,10) rectangle (15,8);

        \node[anchor=north west] at (0.5, 9.75) {\textbf{Assistant Message:}};

        \node[anchor=north west, text width=13.75cm, align=left] at (0.5, 9.25) {Got it! Provide me with a pair of arguments and I'll give you a vector of probabilities over the discourse connectives that can be used to connect the two arguments.};

        \draw (0,8) -- (15,8);

        \fill[new_gray!40] (0,8) rectangle (15,6);

        \node[anchor=north west] at (0.5, 7.75) {\textbf{User Message:}};

        \node[anchor=north west, text width=13.75cm, align=left] at (0.5, 7.25) {Argument 1: Everybody seemed to hear a rustling outside the door. There was no sound of footsteps.\\
        Argument 2: It was like light silk sliding over the panel. Then it stopped.};

        \draw (0,6) -- (15,6);

        \fill[new_gray!0] (0,6) rectangle (15,4);

        \node[anchor=north west] at (0.5, 5.75) {\textbf{Assistant Message:}};

        \node[anchor=north west, text width=13.75cm, align=left] at (0.5, 5.25) {[0.0, 0.0, 0.0, 0.0, 0.1, 0.0, 0.0, 0.0, 0.0, 0.0, 0.0, 0.0, 0.1, 0.1, 0.0, 0.0, 0.0, 0.0, 0.0, 0.0, 0.0, 0.0, 0.0, 0.2, 0.0, 0.0, 0.0, 0.5]};

        \draw (0,4) -- (15,4);

        \fill[new_gray!40] (0,4) rectangle (15,2);

        \node[anchor=north west] at (0.5, 3.75) {\textbf{User Message:}};

        \node[anchor=north west, text width=13.75cm, align=left] at (0.5, 3.25) {Argument 1: It pretends to respect human rights. It pretends to persecute no one.\\
        Argument 2: It pretends to fear nothing. It pretends to pretend nothing.};

        \draw (0,2) -- (15,2);

        \fill[new_gray!0] (0,2) -- (15,2) -- (15,0.1) arc[start angle=0, end angle=-90, radius=0.1cm] -- (0.1,0) arc[start angle=-90, end angle=-180, radius=0.1cm];

        \node[anchor=north west] at (0.5, 1.75) {\textbf{Assistant Message:}};

        \node[anchor=north west, text width=13.75cm, align=left] at (0.5, 1.25) {[0.0, 0.0, 0.0, 0.0, 0.0, 0.0, 0.0, 0.0, 0.0, 0.0, 0.0, 0.0, 0.0, 0.0, 0.3, 0.7, 0.0, 0.0, 0.0, 0.0, 0.0, 0.0, 0.0, 0.0, 0.0, 0.0, 0.0, 0.0]};

    \end{tikzpicture}
    \caption{Prompt-response template for predicting multi-label IDRR in English using DiscoGeM~2.0.}
    \label{fig:prompt}
\end{figure*}

\end{document}